\documentclass[runningheads]{llncs}
\usepackage{graphicx}
\usepackage{CJKutf8}
\usepackage{booktabs}
\usepackage{multirow}
\usepackage{multicol}
\usepackage{tabularx}
\usepackage{color}
\usepackage{amssymb}
\usepackage{amsmath}
\usepackage{siunitx}
\definecolor{orange}{RGB}{255,127,0}
\usepackage{hyperref}
\usepackage{bbding}
\begin{document}
%
\title{RSpell: Retrieval-augmented Framework for Domain Adaptive Chinese Spelling Check}
\titlerunning{RSpell for Domain Adaptive Chinese Spelling Check}
%
\author{Siqi Song\inst{1} \and
Qi Lv\inst{1} \and
Lei Geng\inst{1}  \and
Ziqiang Cao\inst{1,2(}\Envelope\inst{)} \and
Guohong Fu\inst{1,2} 
}
%
%
\institute{School of Computer Science and Technology, Soochow University, China \and
Institute of Artificial Intelligence, Soochow University, China\\
\email{\{sqsong, aopolinqlv, lgeng\}@stu.suda.edu.cn\\
\{zqcao, ghfu\}@suda.edu.cn
}} 
\maketitle              
\begin{abstract}
Chinese Spelling Check (CSC) refers to the detection and correction of spelling errors in Chinese texts. 
In practical application scenarios, it is important to make CSC models have the ability to correct errors across different domains. 
In this paper, we propose a retrieval-augmented spelling check framework called RSpell, which searches corresponding domain terms and incorporates them into CSC models.
Specifically, we employ pinyin fuzzy matching to search for terms, which are combined with the input and fed into the CSC model. 
Then, we introduce an adaptive process control mechanism to dynamically adjust the impact of external knowledge on the model. 
Additionally, we develop an iterative strategy for the RSpell framework to enhance reasoning capabilities.
We conducted experiments on CSC datasets in three domains: law, medicine, and official document writing. The results demonstrate that RSpell achieves state-of-the-art performance in both zero-shot and fine-tuning scenarios, demonstrating the effectiveness of the retrieval-augmented CSC framework. Our code is available at
\url{https://github.com/47777777/Rspell}.
\end{abstract}
\section{Introduction}
Chinese Spelling Check (CSC) aims to detect and correct misspelled characters in Chinese sentences~\cite{2013Integrating}.
It is a fundamental task in natural language processing, widely used in downstream NLP tasks such as speech recognition, summarization, and machine translation.
As mentioned in previous studies~\cite{2010Visually}, almost all Chinese spelling errors are related to phonological and visual similarity.
Therefore, CSC models often integrate grapheme and phonetic information \cite{huang2021phmospell,zhu2022mdcspell}.

Most of the existing CSC research is concentrated in the general area.
Considering practical applications, it is also important for CSC models to have error correction capabilities in different domains.
The mainstream practice is collecting datasets and fine-tuning a specific speller for each domain, which can be less scalable and time-consuming. 
\cite{lv2022general} proposed an unsupervised approach that used domain terms to incorporate relevant knowledge into general spellers. 
Nevertheless, its performance excessively relied on many user-defined hyperparameters.

\begin{CJK}{UTF8}{gbsn}
\begin{table}[tbp]
\caption{Instance of Chinese spelling errors. The \textcolor{red}{wrong}/\textcolor{blue}{golden} characters are in \textcolor{red}{red}/\textcolor{blue}{blue}. In the retrieval phrase, the \textcolor{orange}{orange} indicates words that do not need to be modified in the original sentence, while the \textcolor{green}{green} indicates the correct replacement for the incorrect token in the original sentence.}
  \centering
  \resizebox{\textwidth}{!}{
    \begin{tabular}{c|l}
    \toprule
    \multicolumn{2}{c}{ Instance} \\
    \midrule
    \multirow{2}[2]{*}{Input} & 治疗弱视采用医学验光配镜来进行\textcolor{red}{校}正。 \\
          & The treatment of amblyopia involves the use of medical optometry and corrective lenses for \textcolor{red}{proofreading}. \\
    \midrule
    \multirow{3}[2]{*}{w/ Retrieve} & 治疗弱视采用医学验光配镜来进行\textcolor{red}{校}正。‖领域词是\textcolor{orange}{弱视}，\textcolor{orange}{医学验光}，\textcolor{orange}{配镜}，\textcolor{green}{矫正} \\
          & The treatment of amblyopia involves the use of medical optometry and corrective lenses for \textcolor{red}{proofreading}. \\
          & ‖The field words are \textcolor{orange}{amblyopia}, \textcolor{orange}{medical optometry}, \textcolor{orange}{corrective lenses}, \textcolor{green}{correction} \\
    \midrule
    \multirow{2}[2]{*}{Target} & 治疗弱视采用医学验光配镜来进行\textcolor{blue}{矫}正。 \\
          & The treatment of amblyopia involves the use of medical optometry and corrective lenses for \textcolor{blue}{correction}. \\
    \bottomrule
    \end{tabular}%
  \label{tab1}}%
\end{table}%
\end{CJK}

On many other tasks, such as open-domain question answering~\cite{Chen_Fisch_Weston_Bordes_2017} and machine translation~\cite{Khandelwal_Fan_Jurafsky_Zettlemoyer_Lewis_2020}, retrieval methods are typically used to introduce external knowledge. 
By incorporating external auxiliary information, models are no longer solely reliant on the training corpus and its internal weights, resulting in improved model performance.
Drawing inspiration, we propose a universal retrieval-augmented framework to inject external domain term knowledge into original spellers.
Given an input sentence, we retrieve its relevant domain terms to guide spellers from two aspects:
i). avoiding over correction; ii). mining more potential errors.

Specifically, We first construct domain-specific lexicons that contain Chinese phrases and their corresponding pinyin forms.
we second develop a retriever for Chinese spelling check.
It is difficult to match accurately with external information based on text, given that the given sentence contains misspelled words. 
Most Chinese misspelled tokens are phonetically close to their correctly spelled counterparts, so we transform the sentence into a pinyin string and use fuzzy matching with the pinyin. 
Finally, we concatenate the retrieved domain terms and the original sentence into the speller, as shown in Table~\ref{tab1}.
During training, we propose a process control mechanism to adaptively control the influence of retrieval knowledge. Only when there is a match between the retrieval terms and the target sentence, the retrieval results are incorporated into the speller.
In addition, we design a two-stage retrieval strategy to handle cases where a single sentence has multiple errors.
We evaluate RSpell on the domain-specific CSC dataset~\cite{lv2022general} composed of three domains: law, medicine, and official document writing.
Our results outperform state-of-the-art models in both zero-shot learning and fine-tuning settings, providing strong evidence of the effectiveness of the retrieval-enhanced CSC framework. 

In summary, the main contributions of this paper are as follows:
1) To our knowledge, we are the first to introduce retrieval into CSC tasks.
2) We propose RSpell, a universal framework that can be combined with different spell checkers to enhance their performance on domain-specific data.
3) Our method achieve state-of-the-art results in all three domain-specific datasets, including both fine-tuning and zero-shot scenarios. 

\section{Related Work}

\paragraph{\textbf{Chinese Spelling Check}}
In order to detect and correct spelling errors, early works were mainly based on various manual-designed rules and traditional machine learning methods~\cite{2012A,2014Chinese,Gu2014IntroductionTN,2013Conditional,2015HANSpeller}.
With the rapid development of deep learning, employing pre-trained language models for solving CSC tasks has emerged as a prevailing approach~\cite{wang-etal-2018-hybrid,2019FASPell,cheng2020spellgcn,2020Spelling}.
Researchers found the most important cause of Chinese spelling errors is the similarity of sound and shape.
Thus, a line of studies have incorporated multimodal information into CSC models~\cite{wang2021dynamic,zhang2021correcting,li2022improving,xu2021read,huang2021phmospell}.
Compared with the general domain which the above methods focus on, specific domains are also important for CSC application in practice.
\cite{lv2022general} firstly annotated multi-domain CSC datasets and proposed ECSpell which used lexicons to make the model have the domain-adaptive ability.
Our research also focuses on the domain-related CSC task, introducing the retrieval approach to that task. 

\paragraph{\textbf{Text Information Retrieval}}
Text information retrievers are mainly divided into sparse representation~\cite{2017Reading} and dense representation~\cite{2020Dense}.
The former computes the relevance score according to some specific statistics including TF-IDF~\cite{sparck1972statistical}, BM25~\cite{turnbull2015bm25}.
Since its simplicity and efficiency, many researchers apply it to downstream tasks~\cite{2018Retrieve,DBLP:journals/corr/GuWCL17}.
The latter obtains dense representation from the encoder of a Transformer that has been trained on specific data~\cite{2019Generalization,2020BERT}. 
Therefore, this approach often contains richer and more dense information. 
It is usually applied in open-domain question answering~\cite{Chen_Fisch_Weston_Bordes_2017}.
\cite{2022Interpretability} used examples to improve interpretability by using a dense representation-based retrieval method for the grammar error correction task. 
Motivated by their work, we use sparse representation-based retrieval methods for the CSC task to achieve more accurate error correction.

\section{Our Approach}
\subsection{Problem Formulation}
Given a misspelled sentence $X= \{{ x_1,x_2,\cdots ,x_n}\}$, the CSC model aims to output the correct sentence $Y= \{{ y_1,y_2,\cdots ,y_n}\}$ .
As the sequence lengths of input $X$ and output $Y$ are the same, the CSC task is usually regarded as a token prediction task.
In addition, we define the domain phrases retrieved from the input sentence as $V={\{v_1,v_2,\cdots ,v_m}\}$ to guide the model to make more accurate predictions.

\subsection{Our Framework Overview}
\begin{figure}[htbp]
  \centering
  \includegraphics[width=0.74\linewidth]{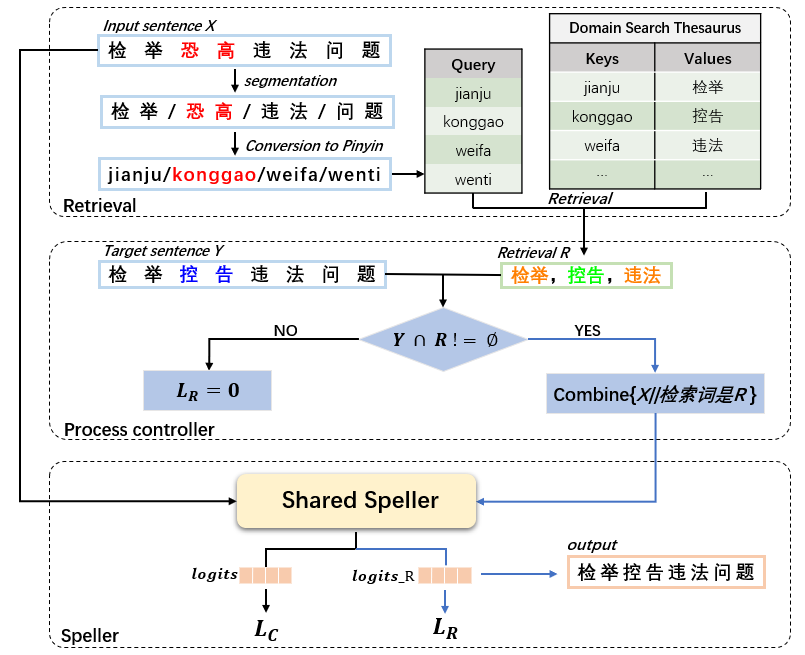}
  \caption{Framework of the proposed RSpell. 
  \textbf{Retrieval:} It retrieves external information relevant to the given sentence from the corresponding domain, i.e., domain phrases with similar pinyin strings. 
  \textbf{Speller:} It denotes a token classification-based speller.
  \textbf{Process controller:} It controls the impact of external information on spellers.}\label{fig:example}
\end{figure}

Our framework uses retrieval techniques to obtain domain knowledge and incorporates it into the speller for more accurate spelling checks. 
The overall overview of the framework is illustrated in Figure~\ref{fig:example}.
We first establish a search engine that utilizes pinyin fuzzy matching to retrieve relevant phrases from the input sentence.
Then, we utilize the retrieved to guide the CSC model, and adaptively control its impact on the model through a process control mechanism. 

\subsection{Retriever For Chinese Spelling Check}
The retrieval-based approach aims to match misspelled tokens in the input with their corresponding correct tokens in the corpus, as well as match correctly spelled tokens in the input with their corresponding correct tokens in the corpus. 
These results are combined with the inputs and feed into the CSC model as prompts.

Given a Chinese sentence $X= \{{ x_1,x_2,\cdots ,x_n}\}$, we first obtain its word sequence according to the off-the-shelf segment tool, Jieba\footnote{\url{https://github.com/fxsjy/jieba}}:
\begin{equation}
    \begin{aligned}
        WORD=\{wd_1, wd_2,\cdots, wd_m\}
    \end{aligned}
\end{equation}
where $wd_i$ denotes the $i$-th word and it may contain one to many characters. $m$ stands for number of $WORD$.

Then, to construct queries of pinyin form, the obtained phrases are converted to its pinyin strings by hanzi2pinyin\footnote{\url{https://pypi.org/project/Pinyin2Hanzi/}}:
\begin{equation}
    \begin{aligned}
        PY=\{py_1, py_2, \cdots, py_m\}
    \end{aligned}
\end{equation}
where $py_i$ represents the corresponding pinyin string of $ph_i$. $m$ stands for number of $PY$.

For each domain dataset, we prepare a dedicated domain thesaurus that contains Chinese phrases and their corresponding pinyin expressions. 
We index the domain lexicon as key-value pairs $ C=(k_i,v_i)$, where the pinyin serves as the key and the corresponding Chinese word serves as the value.
Then, given the input $PY$, each element in $PY$ is treated as a query, and the search engine $Q$ constructed using TF-IDF. A threshold value $\theta$ is set to return the most similar keys and their values.
\begin{equation}
\{(k_1,v_1),\cdots ,(k_r,v_r)\}=Q(PY|C \geq \theta)
\end{equation}
$r$ represents the number of key-value pairs that satisfy the condition of being greater than or equal to $\theta$, $r \leq m$. 
Finally, we obtain the relevant external knowledge $R$ of the original input sentence $X$.
\begin{equation}
R=\{v_1, \cdots, v_r\}
\end{equation}

We concatenate the initial input sentence $X$ and retrieved phrases $R$ with the prompt $p$ to form the final input, where $p$ in our setting is ``\begin{CJK}{UTF8}{gbsn}领域词是\end{CJK}'' which means following phrases are related to the specific domain and also included in the given sentence.
\begin{equation}
X^R=\{x_1,x_2,\cdots ,x_n,p, v_1,\cdots,v_r\}
\end{equation}

\subsection{Adaptive Process Controller}
Due to limitations in retrieval technology and thesaurus capacity, search engines cannot satisfy the need to retrieve all relevant phrases and may introduce some noisy phrases.
In order to balance the effect and noise brought by the retriever, we use an adaptive process control module to control whether the retrieved knowledge is incorporated into the model.

In the training phase, we dynamically judge the retrieved knowledge $R$ and the target sentence $Y$ according to the following condition:
\begin{equation}
    \begin{aligned}
        Condition: R\cap Y\neq \emptyset
    \end{aligned}
\end{equation}
When the $Condition$ is true, which means that there is overlap between the retrieved knowledge $R$ and the text content of the target sentence $Y$, indicating that the retrieved information can effectively help the model and needs to be integrated into the model.
When the $Condition$ is false, which means that there is no similarity between the retrieved knowledge $R$ and the text content of the target sentence $Y$, indicating that the additional information has no effect and does not need to be integrated into the model, $\mathcal{L}_R=0$.

It is worth noting that we only use the adaptive process control module during the training phase. 
In the testing phase, since we do not know the target sentence in advance, we add retrieval information to all sentences to help with error correction, that is, we only follow the right blue branch in Figure~\ref{fig:example} to output the final predicted text.

\subsection{External Knowledge Guided Spell Checker}
The retrieved external knowledge needs to guide the spell checker for error correction. 
It is worth emphasizing that any speller can serve as the speller for RSpell, encoding the original input $X$. 
\begin{equation}
    \begin{aligned}
        E = spellerencoder(X)
    \end{aligned}
\end{equation}
where $E$ is the overall embedding of the original input $X$.
$E \in \mathbb{R}^{l \times 768}$ and $l$ is the length of the input sentence. 
In order to incorporate the retrieved relevant knowledge into the model, We use the encoder of the speller as a shared encoder and input $X^R$:
\begin{equation}
    \begin{aligned}
        E^R = spellerencoder(X^R)
    \end{aligned}
\end{equation}
where $E^R$ is the overall embedding after incorporating external knowledge.
We feed the obtained $E$ and $E^R$ into two layers of Transformers encoders separately to obtain hidden states, $H= \{{ h_1,h_2,\cdots ,h_n}\}$ for $E$, $H^R= \{{ h_1^R,h_2^R,\cdots ,h_n^R}\}$ for $E^R$. 
where $h_i , h_i^R \in \mathbb{R}^{d_t}$ and $d_t$ is the output dimension of the Transformer encoder.
Since the CSC task can be seen as a symbol-level prediction task, the decoder outputs the encoded results and projects them onto a character feature space to predict the correct characters.
\begin{equation}
    \begin{aligned}
        p(\hat{y_i}=y_i|X) =  softmax(W \times h_i)
    \end{aligned}
\end{equation}
\begin{equation}
    \begin{aligned}
        p(\hat{y_i}=y_i|X^R) =  softmax(W \times h_i^R)
    \end{aligned}
\end{equation}
\begin{equation}
    \begin{aligned}
        \mathcal{L}_C =  -\sum_{i=1}^{n}\log p(\hat{y_i}=y_i|X)
    \end{aligned}
\end{equation}
\begin{equation}
    \begin{aligned}
        \mathcal{L}_R =  -\sum_{i=1}^{n}\log p(\hat{y_i}=y_i|X^R)
    \end{aligned}
\end{equation}
where $\mathcal{L}_C$ and $\mathcal{L}_R$ are the character prediction loss and the retrieval knowledge-aided character prediction loss, respectively.
During the training phase, the basic loss for the CSC task needs to be calculated through the left branch. Additionally, by using the adaptive process control mechanism, we determine whether to combine the retrieved content and the original input according to the template and feed it back to the spell checker to calculate the additional retrieval loss, i.e. through the right branch.
The training objective can be summarized as:
\begin{equation}
    \begin{aligned}
        \mathcal{L} = \mathcal{L}_C + \mathcal{L}_R 
    \end{aligned}
\end{equation}

\subsection{Secondary Search Strategy}
We found that the current CSC model performs poorly for sentences with multiple spelling errors. If multiple iterations of error correction are used, the CSC model tends to overcorrect valid expressions into more common ones. 
This issue has also been pointed out by \cite{liu-etal-2022-craspell}. 
Therefore, we propose a second retrieval strategy. Incorporating external retrieval knowledge can not only help the model correct misspellings but also ensure that the model does not alter correct spellings to some extent. 
Based on the result of the first correction, we re-retrieve to obtain more accurate retrieval information and then add the updated knowledge to the error correction model. This method helps to some extent in solving the problem of correcting multiple misspelled words without overcorrecting.

\section{Experiments}
\subsection{Dataset}
\noindent
We use the domain CSC dataset published by~\cite{lv2022general}.
The dataset includes three domains: law, medicine, and official document writing. 
The number of samples in the training and test sets are as follows: law (1960/500), med (3000/500), and odw (1728/500).
For the domain lexicons used in the retrieval module, we use the public Tsinghua University open Chinese dictionaries\footnote{\url{http://thuocl.thunlp.org/}} as the benchmark for the law and medicine fields, and the official document lexicon provided by~\cite{lv2022general} as the benchmark for the official document field. We extract key phrase keywords from relevant corpora using a word segmentation tool and expand the three benchmark lexicons accordingly.

\subsection{Baselines}
\textbf{\textit{BERT}}~\cite{2018BERT} Basic BERT classification model.


\noindent
\textbf{\textit{ReaLiSe}}~\cite{xu2021read} fuses semantic, phonetic and visual information for prediction.

\noindent
\textbf{\textit{SCOPE}}~\cite{li2022improving} introduces the Chinese phonetic prediction assistance task and employs an adaptive weighting scheme to achieve balance.

\noindent
\textbf{\textit{ECSpell}}~\cite{lv2022general} combines glyph information and fine-grained phonetic features, and incorporates a household dictionary-guided inference module.

\noindent
\textbf{\textit{ChatGPT}}\footnote{\url{https://chat.openai.com/}} is an advanced conversational AI model developed by OpenAI.

\noindent
\textbf{\textit{RSpell\textsuperscript{S}}} stands for SCOPE as the speller of RSpell.

\noindent
\textbf{\textit{RSpell\textsuperscript{E}}} stands for ECSpell as the speller of RSpell.

\subsection{Evaluation Metrics and Settings}
We use the evaluation metrics proposed by~\cite{2015Introduction}.
Compared with the character-level metric, the sentence-level metric is more stringent and better tests the performance strength of the model. 
The specific metrics include the precision, recall and F1 score both in of detection and correction level.

Our Framework is based on huggingface's pytorch implementation.
During the training phase, we set the batch to 8, the maximum length sequence to 128, the epoch to 200, the learning rate to 5e-5, and we use AdamW~\cite{loshchilov2017decoupled} as the optimizer.
For zero-shot experiment, we construct general lexicon using the SIGHAN dataset~\cite{2013Chinese,2014Overview,2015Introduction}, and train them using the RSpell framework to activate this retrieval-enhanced capability.
The trained model is then used to directly test on the entire domain data including training and test data.

\begin{table}[htbp]
\caption{RSpell and baselines performance comparison in zero-shot and fine-tuning scenarios across three domains: law, medicine, and official document writing. Best results are in~\textbf{bold}. Testing was performed on the entire dataset in the zero-shot learning scenario, and on the test set in the fine-tuning scenario. }
  \centering
  \resizebox{\textwidth}{!}{
    \begin{tabular}{c|c|ccc|ccc|ccc|ccc}
    \toprule
    \multirow{3}[6]{*}{Dataset} & \multirow{3}[6]{*}{Method} & \multicolumn{6}{c|}{Zero-shot}                & \multicolumn{6}{c}{Fine-tuning} \\
\cmidrule{3-14}          &       & \multicolumn{3}{c|}{Detection Level} & \multicolumn{3}{c|}{Correction Level} & \multicolumn{3}{c|}{Detection Level} & \multicolumn{3}{c}{Correction Level} \\
\cmidrule{3-14}          &       & Pre   & Rec   & F1    & Pre   & Rec   & F1    & Pre   & Rec   & F1    & Pre   & Rec   & F1 \\
    \midrule
    \multirow{6}[4]{*}{law} & BERT  & 76.9  & 65.5  & 70.8  & 69.0  & 58.8  & 63.5  & 82.5  & 77.7  & 80.0  & 76.7  & 72.2  & 74.3  \\
          & ReaLiSe & 48.0  & 45.4  & 46.7  & 35.0  & 33.0  & 34.0  & 69.1  & 67.5  & 68.3  & 63.1  & 61.6  & 62.3  \\
          & SCOPE &51.8  & 58.9  & 55.2  &45.9  & 52.1  & 48.8  & 61.8  & 72.9  & 66.9  & 56.8  & 67.1  & 61.5  \\
       & ECSpell  & 78.2  & 67.8  & 72.6  & 72.2  & 62.6  & 67.2  & 86.1  & 82.4  & 84.2  & 78.3  & 74.9  & 76.6  \\
       & ChatGPT &40.1  & 21.5  & 28.0  & 35.7  & 19.1  & 24.9  &\texttt{-} &\texttt{-} &\texttt{-} &\texttt{-} &\texttt{-} &\texttt{-}   \\
        \cmidrule{2-14}   	 	 	 	 	 
        & \textbf{RSpell\textsuperscript{S}} &55.6   & 63.2  & 59.2  & 48.9  & 55.6  & 52.0  & 67.3  & 74.9  & 70.9  & 61.3  &68.2  & 64.6  \\
       & \textbf{RSpell\textsuperscript{E}} & ~\textbf{80.7 } & ~\textbf{72.5 }  & ~\textbf{76.4 } & ~\textbf{73.5 } & ~\textbf{66.1 } & ~\textbf{69.6 } & ~\textbf{91.0 } & ~\textbf{87.1}  & ~\textbf{89.0}  & ~\textbf{85.3 } & ~\textbf{81.6}  & ~\textbf{83.4 } \\

    \midrule
    \multirow{6}[4]{*}{med} & BERT  & 74.5  & 61.4  & 67.3  & 65.6  & 54.0  & 59.2  & 85.0  & 69.9  & 76.7  & 77.4  & 63.7  & 69.9  \\
          & ReaLiSe & 42.8  & 39.4  & 41.0  & 27.2  & 25.1  & 26.1  & 68.3  & 57.1  & 62.2  & 55.0  & 46.0  & 50.1  \\
          & SCOPE  & 54.2  & 58.0  & 56.1  & 45.9  & 49.1  & 47.4  & 72.0  & 71.7  & 71.8  & 61.3  & 61.1  & 61.2  \\
         & ECSpell  & ~\textbf{75.8 } & 65.8  & ~\textbf{70.4 } & ~\textbf{67.6 } & 58.6  & 62.8  & 84.9  & 79.7  & 82.2  & 75.9  & 71.2  & 73.5  \\
         & ChatGPT &23.5  & 22.2  & 22.8  & 20.4  & 19.3  & 19.9  &\texttt{-} &\texttt{-} &\texttt{-} &\texttt{-} &\texttt{-} &\texttt{-}  \\
         \cmidrule{2-14}  		 	 	 	 
         & \textbf{RSpell\textsuperscript{S}} &  52.0 &62.0  &56.5   & 45.5 & 54.3  & 49.5  & 71.6  & 74.8  & 73.2  & 65.3  & 68.1  & 66.7  \\
        & \textbf{RSpell\textsuperscript{E}} & 73.0  & ~\textbf{67.5 } & 70.1  & 66.7  & ~\textbf{61.7 } & ~\textbf{64.1 } & ~\textbf{89.6}  & ~\textbf{80.1 } & ~\textbf{84.6 } & ~\textbf{86.1}  & ~\textbf{77.0 } & ~\textbf{81.3 } \\

    \midrule
    \multirow{6}[4]{*}{odw} & BERT  & 79.8  & 62.6  & 70.1  & 74.0  & 58.1  & 65.1  & 85.3  & 74.9  & 79.8  & 78.8  & 69.2  & 73.7  \\
          & ReaLiSe & 49.6  & 43.8  & 46.5  & 38.0  & 33.6  & 35.6  & 64.0  & 58.9  & 61.4  & 55.0  & 50.6  & 52.7  \\
          & SCOPE  & 82.6  &73.4  & 77.7  & 75.7  & 67.3  & 71.3  &88.4  & 81.0  & 84.5  & 82.2  & 75.3  & 78.6  \\     
          & ECSpell & 82.4  & 70.1  & 75.8  & 76.9  & 64.3  & 70.2  & 88.2  & 79.9  & 83.8  & 82.3  & 74.5  & 78.2   \\
          & ChatGPT &53.5  & 22.8  & 32.0  & 45.6  & 19.4  & 27.2    &\texttt{-} &\texttt{-} &\texttt{-} &\texttt{-} &\texttt{-} &\texttt{-} \\
          \cmidrule{2-14}   	 	 	 	 	 
          & \textbf{RSpell\textsuperscript{S}}  &84.6   &~\textbf{80.9}   &~\textbf{82.7}   &77.3   &~\textbf{73.9}   &~\textbf{75.5}   & 90.1  & ~\textbf{86.7}  & ~\textbf{88.4}  & 83.4  & ~\textbf{80.2}  &81.8  \\
          & \textbf{RSpell\textsuperscript{E}} & ~\textbf{87.4 } &72.5  & 79.3  & ~\textbf{80.6 } & 66.8  & 73.1  & ~\textbf{92.4 } & 82.9  & 87.4  & ~\textbf{89.0 } & 79.9  & ~\textbf{84.2 } \\
    \bottomrule
    \end{tabular}%
  \label{tab:main_exp}}%
\end{table}%

\vspace{-0.3cm} 
\subsection{Main Results}
Table~\ref{tab:main_exp} shows the sentence-level performance of RSpell and baseline methods on three domain datasets, law, medicine and official document writing, for both zero-shot and fine-tuning scenarios.  
With the help of retrieval framework, RSpell\textsuperscript{S} and RSpell\textsuperscript{E} outperform other baselines.
Compared with the original spellers, our proposed RSpell framework has achieved at least 1.3\% improvement under zero shot setting and 3-8\% improvement under fine-tuning setting, respectively, which indicates that incorporating retrieval information is a highly effective approach for CSC tasks in domain-specific data. 
We can see that the proposed framework achieves a greater improvement under the fine-tuning setting compared to the zero-shot setting. 
We speculate that the most likely reason for this is that the model can better adapt to the new domain-specific data during fine-tuning, which also implies that there is still room for further improvement in the model's generalization ability.
Additionally, we observe that despite the strong language capabilities of large language models, their output format is often unstable, making them less suitable for CSC tasks.

\subsection{Ablation Studies}
To explore the effectiveness of each component of RSpell, we conducted ablation studies with different settings: 
1) removing the information retrieval module (w/o IR), 2) removing the adaptive process controller (w/o APC), and 3) removing the second iteration strategy (w/o SSS). As shown in Table~\ref{tab:ablation_study}, the performance drops regardless of which component is removed, proving the effectiveness of each component.

\begin{table}[htbp]
\caption{Ablation results on test sets in three domains: law, medicine, and official document writing.
The following modifications were made to RSpell\textsuperscript{E}: removing information retrieval (w/o IR), removing adaptive process controller (w/o APC), and remove the secondary search strategy (w/o SSS). It is worth noting that the absence of IR then includes the absence of APC and the absence of SSS. Best results are in~\textbf{bold}.}
  \centering

    \begin{tabular}{l|c|c|c|c|c|c}
    \toprule
    \multicolumn{1}{c|}{\multirow{2}[4]{*}{Method}} & \multicolumn{2}{c|}{Law} & \multicolumn{2}{c|}{Med} & \multicolumn{2}{c}{Odw} \\
\cmidrule{2-7}          & D-F   & C-F   & D-F   & C-F   & D-F   & C-F \\
    \midrule
    \textbf{RSpell\textsuperscript{E}} & ~\textbf{89.0 } & ~\textbf{83.4 } & ~\textbf{84.6 } & ~\textbf{81.3 } & 87.4  & ~\textbf{84.2 } \\
    w/o IR & 84.0  & 76.7  & 79.7  & 74.1  & 82.0  & 74.7  \\
    w/o APC & 87.7  & 82.1  & 83.4  & 78.7  & ~\textbf{87.5}  & 83.5  \\
    w/o SSS & 86.8  & 81.1  & 80.8  & 78.0  & 84.2  & 81.0  \\
    \bottomrule
    \end{tabular}%
  \label{tab:ablation_study}%
\end{table}%

\subsection{Effect of Varying Lexicons Sizes}
Incorporating relevant external knowledge through retrieval can effectively assist the CSC model in error correction, and the degree of the external knowledge's impact is a key factor. 
We found that the size of the retrieval thesaurus has a significant impact on the experimental results. 
We use the word segmentation tool to extract key phrases from the corresponding training corpus and extend the three benchmark lexicons accordingly.
As shown in Table~\ref{tab:thesaurus_exp}, with the increase of the retrieval thesaurus size, the performance improvement becomes more apparent. 
This indicates that expanding the retrieval thesaurus can retrieve more effective information and thus improve the error correction effect.

\begin{table}
\caption{The performance impact of using retrieval thesaurus of different sizes in the fine-tuning scenario was compared on datasets in three domains: law, medicine, and official document writing. Best results are in~\textbf{bold}.}
  \centering
    \begin{tabular}{c|c|ccc|ccc}
    \toprule
    \multirow{2}[4]{*}{Dataset} & \multirow{2}[4]{*}{Lexicons Size} & \multicolumn{3}{c|}{Detection Level} & \multicolumn{3}{c}{Correction Level} \\
\cmidrule{3-8}          &       & Pre   & Rec   & F1    & Pre   & Rec   & F1 \\
    \midrule
    \multirow{2}[2]{*}{law} & w/o Expanding(9896)  & 89.5  & 83.9  & 86.6  & 82.4  & 77.3  & 79.8  \\
          & w/ Expanding(33121) & ~\textbf{91.0 } & ~\textbf{87.1 } & ~\textbf{89.0 } & ~\textbf{85.3 } & ~\textbf{81.6 } & ~\textbf{83.4 } \\
    \midrule
    \multirow{2}[2]{*}{med} & w/o Expanding(18749) & 89.6  & 80.1  & 84.6  & 84.7  & 75.7  & 79.9  \\
          & w/ Expanding(21583) & ~\textbf{89.6 } & ~\textbf{80.1 } & ~\textbf{84.6 } & ~\textbf{86.1 } & ~\textbf{77.0 } & ~\textbf{81.3 } \\
    \midrule
    \multirow{2}[2]{*}{odw} & w/o Expanding(12509) & 92.3  & 82.5  & 87.2  & 88.1  & 78.7  & 83.1  \\
          & w/ Expanding(29778) & ~\textbf{92.4 } & ~\textbf{82.9 } & ~\textbf{87.4 } & ~\textbf{89.0 } & ~\textbf{79.9 } & ~\textbf{84.2 } \\
    \bottomrule
    \end{tabular}%
  \label{tab:thesaurus_exp}%
\end{table}%

\section{Conclusion}
We propose RSpell, a retrieval-augmented framework for domain adaptive CSC. 
RSpell leverages retrieval methods to transform sentences into phonetic sequences based on the characteristics of the CSC task. 
It utilizes fuzzy matching to retrieve relevant external knowledge and guides the spell checker for accurate error correction. 
RSpell sets new benchmarks on three domain-specific CSC datasets, demonstrating that incorporating retrieval information is a highly effective approach for CSC tasks.

\section*{Acknowledgments}
We thank all reviewers for their valuable comments. 
This work was supported by the Young Scientists Fund of the National Natural Science Foundation of China (No. 62106165), 
the National Natural Science Foundation of China (No. 62076173).


%
%
%
\bibliographystyle{splncs04}
\bibliography{mybibliography}
%





\end{document}